# Autonomous Vehicles Meet the Physical World: RSS, Variability, Uncertainty, and Proving Safety

Philip Koopman✉, Beth Osyk, Jack Weast

Edge Case Research, Pittsburgh PA, USA
Intel, Chandler, AZ, USA
koopman@cmu.edu, bosyk@ecr.guru, jack.weast@intel.com

**Abstract.** The Responsibility-Sensitive Safety (RSS) model offers provable safety for vehicle behaviors such as minimum safe following distance. However, handling worst-case variability and uncertainty may significantly lower vehicle permissiveness, and in some situations safety cannot be guaranteed. Digging deeper into Newtonian mechanics, we identify complications that result from considering vehicle status, road geometry and environmental parameters. An especially challenging situation occurs if these parameters change during the course of a collision avoidance maneuver such as hard braking. As part of our analysis, we expand the original RSS following distance equation to account for edge cases involving potential collisions mid-way through a braking process.

We additionally propose a Micro-Operational Design Domain (μODD) approach to subdividing the operational space as a way of improving permissiveness. Confining probabilistic aspects of safety to μODD transitions permits proving safety (when possible) under the assumption that the system has transitioned to the correct μODD for the situation. Each μODD can additionally be used to encode system fault responses, take credit for advisory information (e.g., from vehicle-to-vehicle communication), and anticipate likely emergent situations.

## 1 Introduction

The Responsibility-Sensitive Safety (RSS) model proposes a way to prove the safety of self-driving vehicles [Shalev-Shwartz17]. The RSS approach is currently deployed in Intel/Mobileye's test fleet of fully automated vehicles. Application areas of RSS include both fully autonomous vehicles and driver assistance systems. This paper reports results of an ongoing joint project to externally validate and further improve RSS.

A salient feature of RSS is the use of Newtonian mechanics to specify behavioral constraints such as determining safe following distance to avoid collisions even when other vehicles make extreme maneuvers such as hard braking. Employing RSS as safety checking logic requires not only knowledge of the physics of the situation, but also correct measurements to feed into the RSS equations.

We consider an example of applying RSS rules to a longitudinal following distance scenario involving the vehicle under consideration (often called the *ego vehicle*) as a follower behind a lead vehicle. To put RSS into practice, the ego vehicle requires at least some knowledge of the physical parameters fed into the physics equations, including ego vehicle and lead vehicle status, road geometry, and operational environment. However, proving guaranteed safety via that approach is complicated by variability and uncertainty. Also worrisome are environmental situations that can transition the system into an unsafe situation even though each vehicle was following RSS safety rules.

This paper identifies the implications for these issues in applying RSS to real vehicles. Additionally, it proposes a new following distance equation to encompass edge cases that were out of scope for the original RSS analysis.

A significant finding is that variability and uncertainty in the operational conditions introduce significant challenges for ensuring safety while maintaining acceptable permissiveness. (The permissiveness of a system is how free it is to operate without violating safety constraints [Guiochet08].) Variability is especially problematic because of the large potential dynamic range of driving conditions. For example, the difference between safe following distance on an icy hill compared to flat dry pavement means that a one-size-fits-all worst case approach to safe following distance is unlikely to result in a vehicle people will actually want to use. This paper seeks to identify the issues that must be resolved to use the RSS equations in a way that provides provable safety to the maximum degree practicable.

In considering how to solve challenges arising from variability and uncertainty we avoid an approach of feeding statistical distributions through the RSS equations. That is because such an approach, as well as other probabilistic analysis approaches (e.g., [Laugier11]) could negate much of the benefit of having a formally provable aspect of safety. By the same token, arbitrarily bounding values to “worst” cases that are only probabilistically true simply buries the problem.

Instead, we separate stochastic aspects of the system into a mode transition approach among Operational Design Domain (ODD) [NHTSA17] segments. Each strictly bounded Micro-ODD (μODD) region can be proven safe. Uncertainty manifests as the probability that the μODD selected actually encompasses the current real-world operational environment. This supports a Bayesian approach to selecting a context-dependent operational posture to maximize permissiveness while supporting cautious behavior in higher risk situations. Further, it allows for an extended range of operational conditions, including behavioral responses to fault scenarios and behavioral responses that take advantage of potentially better knowledge such as vehicle-to-vehicle communication.

## 2 Related Work

Advanced Driver Assistance Systems (ADAS) have made large strides in improving automotive safety, especially in mitigating the risk of rear end collisions. Autonomous Emergency Braking (AEB) can now fully stop a vehicle in many lower-speed situations [ADAC13]. Beyond AEB, vehicles may offer driver assistance technologies including a safe distance warning [ADAC13]. Technologies have differing availability depending on speed and manufacturer [Hulshof13]. Test protocols generally select a few speed combinations representative of urban and highway driving [Hulshof13] in controlled conditions. Moreover, it is typical for current ADAS systems to used fixed rules of thumb (e.g., the two-second following rule as used by [Fairclough97]) for establishing operational safety envelopes that while potentially improving safety on average can either be to conservative or too optimistic. This paper takes a broader approach that considers the specifics of the vehicles involved and environmental conditions. We are not aware of other work that considers expanding physics-based safety analysis such as RSS to consider environmental conditions and vehicle performance characteristics.

Concern about covering all operational conditions is included in upcoming vehicle standards. The Safety of the Intended Functionality (SOTIF) standard treats the universe as partitioned into four scenario sets: known safe, known unsafe, unknown safe, and unknown unsafe [SOTIF17]. The goal is to maximize the known safe area by minimizing the known unsafe area and discovering as many new unsafe scenarios as possible with a given level of effort [SOTIF17]. RSS with the μODD approach can help formalize and enlarge the known safe set while achieving acceptable permissiveness.

Another approach to ensuring autonomy safety is to explain the behavior of control systems. For neural networks, new techniques offer the ability to visualize activations [Yosinski15], and offer various ways to reconstruct an image based on network activations and features [Mahendran16]. Proposals use various names for competency expression, such as communicating self-confidence in a task [Sweet16], or cognitive autonomy where an autonomy system should possess "reasoning where it can introspect about its own reasoning limitations and capacity" [Mani18]. While these approaches might eventually be widely deployed, an RSS-like approach can use a safety envelope approach to ensure safety even if autonomy components have not been fully validated.

To eventually reach Full Driving Automation Level 5 [SAE18] (or even highly permissive Level 4 capabilities), vehicles will need to seamlessly cover a wide range of driving scenarios and environmental conditions. Current work in support of this goal includes taking an ontology-based approach to categorizing driving scenes (e.g., in the PEGASUS project [Bagschik18]). While a run-time vehicle safety checker can benefit from such an ontology, it is additionally helpful to concentrate design effort on the most important scenarios while under-approximating ideal permissiveness for uncommon scenarios to reduce complexity and cost. The RSS model combined with the μODD approach present a strategy for covering the range of operational conditions plus potential for guiding test point selection based on μODD boundaries that might differ from collected design scenarios and functionality test plans. A specific strength of our approach is that it provides a way to handle "unknowns," including uncertainty as to which scene is actually present in the external world, gaps in the driving scene ontology, and what to do when the vehicle is forcibly thrown out of its ODD by external events.

Work on characterizing and dealing with perception uncertainty in the context of safety critical systems is still developing. [Czarnecki18] provides a model of factors that influence development and operational uncertainty.

Safe state analysis is a theme for autonomous vehicle path planning. Path planning algorithms may consider the safety of the current state and reachable states in order to plan a path, including making predictions about potentially occluded obstacles [Orzechowski18]. Such approaches tend to suffer from probabilistic limitations on their ability to provide deterministic safety, whereas the RSS approach to safety aspires to provide a deterministic model for safety. The μODD approach additionally examines the problem of improving the permissiveness of a deterministic safety checking function in the face of variability and uncertainty about the environment.

We base our analysis on an initial RSS paper [Shalev-Shwartz17], and are aware of a follow up paper [Shalev-Shwartz18]. Interest in the performance aspect of RSS continues to grow, with a model and analysis of traffic throughput presented in [Mattas19] comparing RSS to human drivers under various values for the RSS parameters. We are not aware of other published analyses of RSS equations for correctness and completeness.

# 3 RSS Overview

## 3.1 The RSS Following Distance Equation

In an RSS leader/follower scenario, the follower vehicle is presumed to be responsible for ensuring a safe longitudinal distance, so we assume that the ego vehicle is the follower. For this situation, RSS uses a safety principle of: "keep a safe distance from the car in front of you, so that if it will brake abruptly you will be able to stop in time." [Shalev-Shwartz17] Figure 1 shows a notional vehicle geometry:

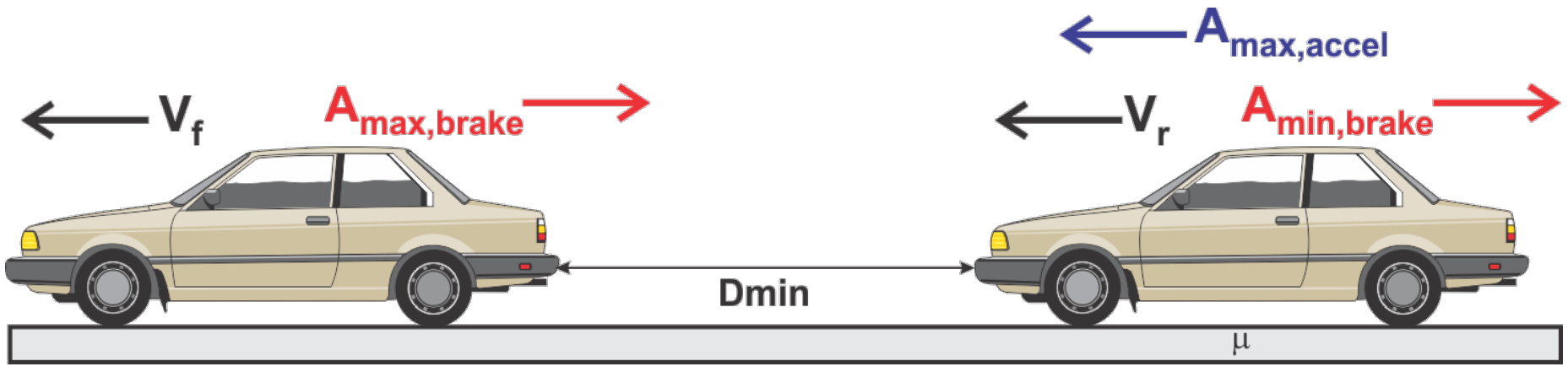

Figure 1. Reference vehicle geometry for leader/follower.

This yields a minimum following distance (id., Lemma 2):

$$d'_{min} = MAX\left\{0, \left(v_r\rho + \frac{1}{2}\, a_{max,accel}\rho^2 + \frac{(v_r + \rho a_{max,accel})^2}{2a_{min,brake}} - \frac{v_f^2}{2a_{max,brake}}\right)\right\} \quad (1)$$

Where in our case the ego vehicle is the following ("rear") vehicle, and:

- $d'_{min}$ is the minimum following distance between the two vehicles for RSS
- $v_f$ is the longitudinal velocity of the lead ("front") vehicle
- $v_r$ is the longitudinal velocity of the following ("rear") vehicle
- ρ is the response time delay before the ego (rear) vehicle starts braking
- $a_{max,brake}$ is the maximum braking capability of the front vehicle
- $a_{max,accel}$ is the maximum acceleration of the ego (rear) vehicle
- $a_{min,brake}$ is the minimum braking capability of the ego (rear) vehicle

The $d'_{min}$ equation considers a leading vehicle, going at initial speed $v_{f,}$ which executes a panic stop at maximum possible braking force $a_{max,brake}$. The ego following vehicle traveling at $v_r$ is initially no closer than distance $d'_{min}$. In the worst case, the ego vehicle is accelerating at $a_{max,accel}$ when the lead vehicle starts braking. There is a response time ρ during which the ego vehicle is still accelerating. Then the ego vehicle detects the lead vehicle braking and reacts by panic braking with deceleration of at least $a_{min,brake}$. RSS considers the worst case scenario to be a highly capable lead vehicle with high $a_{max,brake}$ followed by an ego vehicle that brakes at an initially lower braking capability of at least $a_{min,brake}$. A poorly braking follower requires additional distance to accommodate its inability to stop quickly.

While a derivation based on comparative stopping distances confirmed the equation, the modeling for Figure 4 (discussed later) and analysis using Ptolemy II [PtolemyII] revealed edge cases beyond the scope of the analysis in [Shalev-Shwartz17]. (Additional RSS braking profile information is provided by [Shalev-Shwartz18].) Specifically, Eqn. 1 does not detect situations in which the two vehicle positions overlap in space during – but not at the end of – the braking response scenario.

As a thought experiment, consider an ego vehicle with good brakes that has matched speeds with a leader of significantly worse braking ability. Eqn. 1 is derived assuming the minimum vehicle separation occurs at the final rest positions. If the rear vehicle has superior braking, it could mathematically be "ahead" of the lead vehicle at some time during braking, yet still have a final rest position "behind" the lead vehicle due to shorter stopping distance. In reality, this is a crash. Thus, an additional constraint is that the rear vehicle must remain behind the lead vehicle at all points in time.

A related scenario is a rear vehicle approaching with high relative velocity and superior braking. The rear vehicle might collide during the interval in which both vehicles are braking, while still having a computed stopping point behind the lead vehicle.

To address these situations, we break the analysis up into two parts based on the situation at the time of a collision if following distance is violated: (1) impact during response time ρ and (2) impact after ρ but before or simultaneous with the rear vehicle stopping. (Impact is no longer possible after the rear vehicle stops for this scenario.)

Accounting for situation (1) requires computing the distance change during the response time ρ. There are two cases. The first is when the front vehicle stops before ρ, and the second is when the front vehicle stops at or after ρ.

Situation (2) has two parts. First, compute the distance change during ρ:

$$d''_{min} = (v_r - v_f)\rho + \frac{(a_{max,accel} + a_{max,brake})\rho^2}{2} \quad (2)$$

Next, solve for the distance between the two vehicles after ρ as a function of time:

$$d'''_{min} = \left(v_r + a_{max,accel}\rho\right)t_r - \frac{a_{min,brake}t_r^2}{2} - \left(\left(v_f - a_{max,brake}\rho\right)t_f - \frac{a_{max,brake}t_f^2}{2}\right) \quad (3)$$

This is a parametric equation involving the time after the response time for both vehicles: $t_r$ ant $t_f$. The minimum distance will occur at time $t_r$=$t_f$=t when both vehicles have equal speed (with the value of t then substituted into Eqn. 3 for evaluation):

$$t = \frac{(v_{r0} - v_{f0}) + (a_{max,accel} + a_{max,brake})\rho}{(a_{min,brake} - a_{max,brake})} \quad (4)$$

The special case minimum following distance is the sum of $d'_{min}$ and $d''_{min}$, and only holds when the rear vehicle is faster than the front vehicle at the end of the response time *and* the rear vehicle can brake better than the front vehicle:

$$d_{min} = \begin{cases} MAX[d'_{min}, (d''_{min} + d'''_{min})] & ;\ special\ case \\ d'_{min} & ;\ otherwise\ (Original\ RSS) \end{cases} \quad (5)$$

Because $a_{max,accel}$ is likely to be of secondary importance for small ρ, we focus the balance of our discussion on braking. However, similar issues apply to acceleration.

### 3.2 Coefficient of Friction

Implicit in the RSS equations is that the maximum frictional force exerted by the vehicle on the ground limits braking ability ([Walker08] pg. 119):

$$F_{friction} = \mu * F_{normal} \quad (6)$$

where:

- $F_{friction}$ is the force of friction exerted by the tires against the roadway
- μ is the coefficient of friction, which can vary for each tire
- $F_{normal}$ is the force with which the vehicle presses itself onto the road surface

The friction coefficient is a property of both the tires and the road surface. It is important to note that μ can be above 1.0 for some materials ([Walker08] pg. 119), so a rigorous proof cannot assume limited μ without placing constraints upon installed tires.

### 3.3 The Normal Force and Road Slope

The normal force on each tire is a property of the vehicle weight, weight distribution, the effects of suspension, the slope of the road, and so on. The normal force is the weight of the vehicle multiplied by the cosine of the road slope, shown by Figure 2:

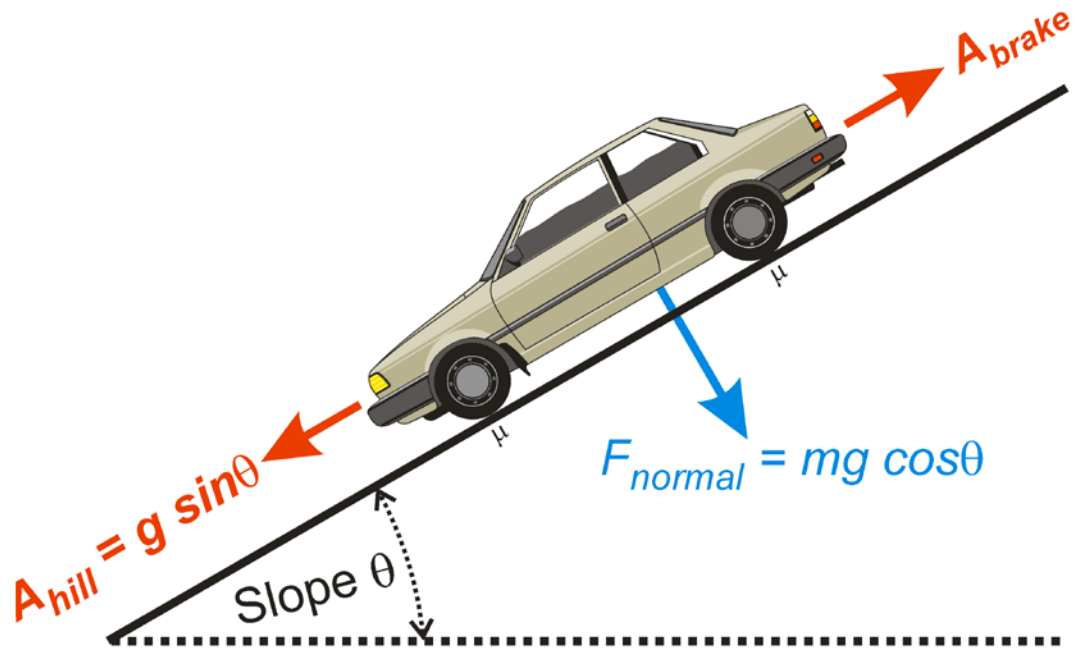

Figure 2: Vehicle forces on an inclined roadway.

In this situation, braking ability is potentially limited by the reduced normal force. Moreover, gravity is pulling the vehicle down the hill, acting against and further reducing the net braking force ([Walker08] pg. 102). If μ is low, the net force can result in the vehicle sliding down the hill (either forwards or backwards) if the brakes cannot overcome the gravitational downhill force vector. Transverse road slope (camber) can similarly reduce $F_{normal}$, but at least does not affect vehicle speed directly.

### 3.4 Road Curvature

An additional limitation to braking capability is that the centripetal force exerted by a vehicle to make turns must be provided by $F_{friction}$ ([Walker08] pg. 128). The net vehicle acceleration (both radial and linear) is a result of a force vector applied by the tire contact patches to the road surface. It follows that any force used to curve the vehicle trajectory steals available force from the ability to stop the vehicle by requiring a force vector that is at off-axis from the vehicle's direction of travel. That means that if the ego vehicle is in a tight turn it will have trouble braking effectively. Lane positioning and racing line techniques [Kapania16] add additional complexity.

A banked curve complicates analysis even further, involving potential increases or decreases to $F_{normal}$ depending upon whether the bank (superelevation) is tilted toward or away from the center of the curve. Secondary factors such as vehicle suspension, tire deformation, transfer of momentum to road debris such as gravel, and so on add additional layers of complexity to any attempted analysis.

## 4 Uncertainty and Variability

While Newtonian Mechanics provides us the tools to determine following distance in principle, even a simplified equation setup for a vehicle's maximum stopping distance on a downhill corkscrew turn is worthy of a college Physics final exam. But in the real world we don't actually know the precise values of all the variables in the equations.

An important issue with proving safety in a cyber-physical system is that there is inherent uncertainty in sensor measurements. That uncertainty includes both issues of accuracy (how close the measurement is to the actual value being measured) and precision (what the distribution of errors in the measurement is across multiple measurements). Uncertainty can additionally be characterized as aleatory uncertainty (e.g., sensor noise that causes non-zero precision), and epistemic uncertainty (e.g., inaccurate measurements and incorrect modeling of the environment) [Chen18]. Both types of uncertainty impair the ability to formally prove safety for a real-world system.

The mere existence of a probability distribution for aleatory uncertainty impairs the ability to create a perfect proof. In principle any series of data points might, with some probability, be wildly inaccurate. Data filtering and statistical techniques might improve the situation, but in the end there is always some non-zero (if infinitesimal) probability that a string of outlier data samples will cause a mishap. Over-sampling to drive that uncertainty below life-critical confidence thresholds (e.g., failure rate of $10^{-9}$/hr) could be impracticable due to the fast time constants required for vehicle control.

For epistemic uncertainty, a significant problem is providing a completely accurate model of the environment and the vehicle. Moreover, even if limitations on sensors and potential correlated sensor failures are mitigated through the use of high-definition maps, variability of operational environments is a significant issue.

Uncertainty cannot be completely eliminated in the real world, so the question is how to account for it within the RSS model while keeping the system practical and affordable. In support of that, we consider sources of uncertainty and variability.

### 4.1 Other Vehicle Parameters

Ensuring that the ego vehicle avoids colliding with other vehicles requires understanding the state of those other vehicles. Knowing where they are and where they are going requires other vehicle pose and kinematic information: {position, orientation, speed, acceleration, curvature} in addition to a prediction of how that information is going to change in the near future (e.g., path plan). That information will be imperfect.

In the absence of perfect information, RSS simply assumes that distance is known and that the lead vehicle will immediately execute a panic braking maneuver at $a_{max,brake}$. While in an ideal world all vehicles have a predetermined and consistent $a_{max,brake}$, in the current world not all vehicles are thus equipped. However, even if new vehicles are standardized, braking capability can increase further due to factors such as after-market brake upgrades, after-market tire upgrades, low tire pressure, after-market aerodynamic modifications, and even driver leg strength. While a vehicle might be equipped with a feature that intentionally limits maximum deceleration, too strict a limit would extend stopping distance and increase collision rates in other situations such as single car crashes.

If the ego vehicle wants to optimize following distance based on the actual lead vehicle capabilities, it will need a way to determine what those are. Most vehicles are not designed to brake above 1g, but it is likely this limit is not universal on public roads. In other words, an assumption of any particular limit for maximum braking capability might be true most of the time, but perhaps not all the time.

### 4.2 Ego Vehicle Parameters

While knowing the exact state of the lead vehicle is difficult, it is also important to appreciate that knowing the state of the ego vehicle is also difficult. Many of the parameters that affect the lead vehicle also affect the ego vehicle, although the concern in this case is more about unexpectedly reduced braking ability. Some factors that might reduce braking capability below expectations include:

- Transient equipment degradation: brake fade due to overheating, brake wetness (e.g., due to puddle splash), cold tire temperature, etc.
- Equipment condition: brake wear, brake actuator damage, low tire tread depth, high tire pressure, etc.
- System interactions: interactions between braking system and electronic stability control, effect of anti-lock braking features, etc.

### 4.3 Environmental Parameters

Successfully executing an aggressive braking maneuver involves not only the vehicle, but also the environment. While environmental conditions in a road segment might be reasonably well known via a local weather service (which becomes safety critical as soon as it is relied upon for this purpose), average values might differ substantially from the instantaneous environmental conditions relevant to a braking maneuver. After all, it is not the average road conditions over a kilometer of road that matter, but rather the specific road conditions that apply to paths of the set of tire contact patches of each vehicle during the course of a panic stop maneuver. Relevant factors that could result in a faster-than-expected lead vehicle braking maneuver combined with a slower-than-expected ego vehicle braking maneuver due to differences on the roadway include:

- Road surface friction: road surface, temperature, wetness, iciness, texture (e.g., milled ridges that increase traction; bumps that cause loss of tire contact), etc.
- Road geometry: slope, banking, camber, curvature as previously discussed
- Other conditions: hydroplaning, mudslides, flooding, high winds pushing against a high profile vehicle body, road debris, potholes, road buckling, etc.

While the two vehicles will traverse the same stretch of roadway for some braking time, their contact patches are not necessarily going to follow exactly the same paths. Localized tire track road conditions can result in different stopping ability even if we attempt to measure some average value of $\mu$. Consider, for example, a lead vehicle that brakes hard in snowy weather on a cleared tire path while the following ego vehicle gets caught slightly laterally displaced from the tracks with its tires on ice.

One may suggest selecting conservative values for these assumptions so as to provide a safety buffer in the estimations. While this may indeed reduce the risk of a crash, our goal is to optimize the permissiveness of the automated driving system to maximize usefulness and infrastructure utilization. Overly conservative values to protect against an incorrect assumption could result in an automated vehicle that may be safe, but may also be quite annoying on the road due to excessively timid vehicle operation.

### 4.4 Potential Assumption-Violating Actions

Even if we know the values for all the variables, there are assumptions made by the RSS longitudinal safety guarantees and stated scope limitations that might be violated by real world situations. Examples include:

- Lead vehicle does not involuntarily move backwards (e.g., due to an icy hill [YouTube17a]; this situation is generally covered by the safe longitudinal distance opposite direction case in RSS [Shalev-Shwartz17].)
- Lead vehicle does not violate the assumed maximum braking deceleration limit (e.g., due to impact with a large boulder that suddenly falls onto the road).
- Roadway $\mu$ does not unexpectedly change (e.g., flash ice-over, or a cargo spill such as a slime eel spill [YouTube17b]).
- Ego vehicle does not fall below minimum expected braking capability (e.g., due to brake fade, puddle splashes onto brake rotor).

- There are no significant equipment failures (e.g., catastrophic brake failure of ego vehicle during a panic braking event).
- There are no unusual vehicle maneuvers (e.g., cut-in scenarios in which a vehicle suddenly appears too close; cut-out scenarios in which the lead vehicle swerves to reveal a much slower, too-close new lead vehicle [EuroNCAP17]).

### 4.5 Effects of Uncertainty and Variability

The concept of an ODD generally corresponds to the notion of the intended operational environment of a vehicle. While it is sometimes equated to a notion of geo-fencing, the dimensions of an ODD are numerous and varied [Koopman19]. As the scope of a planned deployment grows, the cross-product of all possible values for all possible aspects of the ODD suffers a combinatorial explosion, and taking a global worst case can get very pessimistic indeed.

To illustrate how bad a global worst case might be, consider a steep valley with roads covered in ice near the valley bottom. The ego vehicle starts a descent on dry pavement, then hits ice halfway down a steep grade, sliding all the way to the bottom with no ability to stop. Meanwhile, the lead vehicle has already slid down the hill and back up the other side with essentially no ability to stop. The ego vehicle had left a large following distance – which in the end was insufficient due to the lead vehicle sliding backward down the far side of the valley to collide with the ego vehicle at the bottom.

Perhaps this sounds extreme. But cars and even trucks do in fact slide backwards down icy hills on occasion. The point is that most of the time the world is not displaying worst case conditions. It is highly desirable to exploit favorable cases to optimize performance while still making allowances for the worst case. Leaving a kilometer following distance between cars to handle a worst case icy hill is unlikely to be tolerated by other road users on a warm summer day.

## 5 Dealing with Uncertainty and Variability in RSS

In the end, uncertainty can't be eliminated in a system operating in the real world. Rather, we recommend embracing the uncertainty by sandboxing formal proofs of safety into defined μODD segments. Uncertainties regarding the system can then be taken care of by a mechanism for selecting which μODD is currently active.

### 5.1 A Divide-and-Conquer Strategy

We propose dividing the ODD for a vehicle deployment into numerous different μODD regions, with each μODD being selected to have a useful bound on various parameters relevant to RSS safety. For example, some μODDs will correspond to icy road surfaces with steep hills, and others to dry pavement on flat ground. Each μODD can then be evaluated for the worst case following distance required ***for that region's bounded parameters*** rather than for all possible values of parameters across the entire ODD.

While it might seem at first glance that a mathematic approach using bounded ranges of values would suffice, μODDs have two additional advantages. The first is that each segment can be mapped onto a concrete validation test that physically exercises the mathematically computed worst case within each μODD's bounds. (This might or might not correspond to an ontology-based test used for functional validation, depending upon the μODD approach taken.) This can directly tie simulation and physical testing to the underlying representation of safety enforcement.

The second advantage is that μODDs can encode not only geographic location and weather, but also other aspects of the situation including the set of obstacles and events that are expected to be encountered, set of own-vehicle maneuvers that will be intentionally performed, and faults that are expected (e.g., as listed in [Koopman19]).

As shown in Figure 3, all possible system states within the ODD must be accounted for, with RSS used to compute a safety envelope for each μODD. Areas outside the defined ODD can be designated as one or more defensive regions in which the vehicle does its best to remain safe despite having been forced out of its defined overall ODD. In most cases the defensive μODD would invoke a safe maneuvering behavior such as bring the vehicle to a stop. However, some μODDs could also be designed to invoke more specific safing behaviors for pre-identified failure modes and other normal operation ODD excursions. Some smaller μODDs can be optimized for narrow but common special cases (e.g., 54-56 mph on a relatively flat dry divided highway). Other larger generic μODDs can be less optimized in exchange for reduced engineering effort (e.g., ice patches on a hill are likely to result in a cautious vehicle regardless of precise hill slope and posted speed limit).

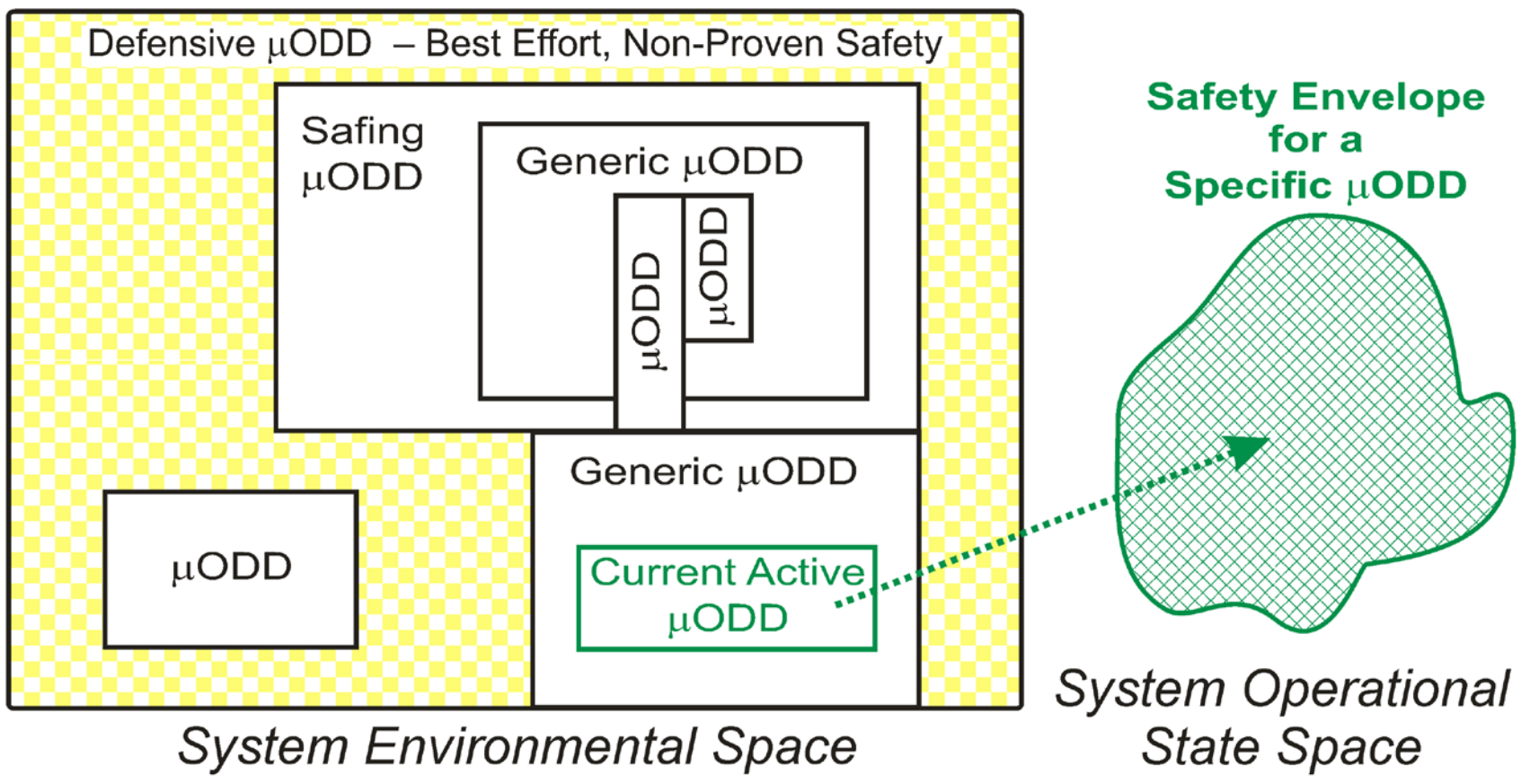

Figure 3. ODD subdivision and mapping onto safety envelopes.

A way to implement a system using this approach is with an ego vehicle ODD mode state machine that transitions between μODD mode states depending upon estimated environmental conditions. An advantage to this approach is that it permits emphasizing engineering effort on the most-used and most important parts of the system operational space (e.g., typical road conditions) while being conservatively safe in unusual operational conditions that aren't worthy of detailed engineering optimization effort.

### 5.2 Partitioning Criteria and Example

Because each μODD has precisely specified boundaries, relevant aspects of safety within each μODD can be proven using RSS. Obviously this approach does not make uncertainty go away. But, what it does provide is a clean separation between provable system properties (RSS-based safety within an active μODD) and uncertainty (whether we have selected the correct μODD, which amounts to whether the actual environmental conditions fall within the specified bounds of the currently active μODD selection).

While at first blush this approach might seem like it is just shuffling uncertainty around, we believe that it re-casts uncertainty into a form more amenable to application to real-world driving situations. That is because it accommodates both measurement uncertainty and modelling uncertainty such as prediction uncertainty.

Knowing that there are many vehicle and environmental parameters that can affect stopping ability, which parameter variations should inhabit separate μODDs and which should belong to the same μODD's equivalence class? Here are some factors we think will be relevant:

**Effect on permissiveness:** Fundamentally, the goal of μODDs is to preserve permissiveness where possible, meaning in this case optimizing for minimal $d_{min}$. Splitting a large μODD into multiple smaller ones is only useful if doing so will enable a substantively smaller $d_{min}$ to be used for a commonly occurring real-world situation.

**Temporal stability:** Some μODDs should be designed to assume relatively constant conditions, while others could conservatively anticipate possible change. For example, flat dry pavement likely to stay flat and dry based on map and weather data suggests creating a μODD accordingly. However, there should be a μODD accounting for the possibility of ice patches. (Whether this should be a conservative pure ice μODD or a more aggressive heterogeneous ice patch μODD is an engineering tradeoff – so long as there is some μODD defined that can cope with worst case ice.) The system may also pre-emptively switch before expected future changes, such as slowing down prior to driving over a bridge in near-freezing weather based on road map data and dropping temperature. Temporal variations can also include contextual information relevant to cautiousness. For example, if a World Cup match has just ended, the μODD could reflect an increased cautiousness for possible occluded pedestrians in the area surrounding the stadium.

**Observability:** Separate μODDs are useful only when differing parameter values are generally observable or can be estimated with sufficiently low uncertainty. This might require vehicle activities to self-characterize, such as exercising brakes occasionally to check that braking capability has not changed. (Significantly more work might need to be done in this area to achieve an appropriate balance of self-test overhead and instrumentation cost.)

**Accounting for external information:** Defining μODDs that exploit external data such as data transmitted from infrastructure and other vehicles is likely to help optimize performance. Whether that data is valid has some uncertainty that should be dealt with by the μODD state transition logic.

Figure 4 shows an illustrative example of μODD decomposition. In this case the system designer has elected to partition the ODD into non-uniform segments based on estimated $a_{min,brake}$, $a_{max,brake}$ values, with an expectation that narrower ranges are more common in practice. Each table cell has a $d_{min}$ value computed based on the worst case for that cell. As can be seen, the safe following distance of more than half a kilometer for the ego vehicle on ice with a lead vehicle on a good road surface is impractical to enforce under more ordinary conditions. (In a practical system there would be multi-dimensional partitioning for bins of $v_r$, $v_f$, and potentially other factors as well.)

| | a_min,brake (g) | | | | | | |
|---|---|---|---|---|---|---|---|
| a_max,brake (g) | 0.05-0.1g | 0.1-0.3g | 0.3-0.4g | 0.4-0.5g | 0.5-0.6g | 0.6-1.0g | 1g+ |
| 0-0.3g | 621.0 | 263.8 | 25.7 | **5.2** | **2.9** | **2.2** | **1.4** |
| 0.3-0.5g | 663.5 | 306.3 | 68.2 | 38.4 | 20.6 | **8.8** | **2.6** |
| 0.5-0.6g | 674.1 | 316.9 | 78.8 | 49.1 | 31.2 | 19.3 | **3.6** |
| 0.6-0.7g | 681.7 | 324.5 | 86.4 | 56.6 | 38.8 | 26.9 | **5.3** |
| 0.7-1.0g | 695.3 | 338.2 | 100.1 | 70.3 | 52.4 | 40.5 | 16.7 |
| 1g+ | 727.2 | 370.0 | 131.9 | 102.2 | 84.3 | 72.4 | 48.6 |

Figure 4. Example $d_{min}$ in meters for example μODDs with varied $a_{max,accel}$ and $a_{min,brake}$ values.
Conditions: $v_r$=25, $v_f$=25, $a_{max,accel}$=0.3g, ρ=0.5 sec.
Cells highlighted where the special case $d_{min}$ prevails.

Implementing a μODD safety feature according to Figure 4 would involve setting up a state machine that transitions between different cells in the table as conditions change. (Note that implicit in this table is an estimate of coefficient of friction that limits braking ability.)

### 5.3 A Bayesian Approach To ODD State Transitions

A purely mathematical approach using worst case bounding values might have been used instead of the μODD table illustrated in Figure 4 if all we cared about was braking force. However, the μODD technique additionally provides a way to account for prior expectations in the state transition engine.

Consider, for example, the question of how fast to go down an urban street next to a row of parked cars. There is a non-zero chance that a child will dart out into traffic, making it prudent to go below the posted speed limit in high risk situations. But if we treat every parked car as certain to be harboring a child waiting for the worst possible moment to jump out into the roadway, we'd have gridlock 24 hours a day. Instead, human drivers use their judgment (albeit imperfectly) based on context.

We can formalize this approach using μODDs by incorporating Bayesian network-style prior belief of risk into the state transition arcs. Figure 5 shows a simplified example to give an idea of the principle involved. The idea is that the Bayesian prior belief of a child running into a roadway is low in the middle of the night, but medium if children are nearby in general. However, a general rule of thumb is that every ball rolling into the street is likely to be followed by a heedless child (triggering a high prior belief of an emergent child), resulting in a transition to extremely low speed permissiveness.

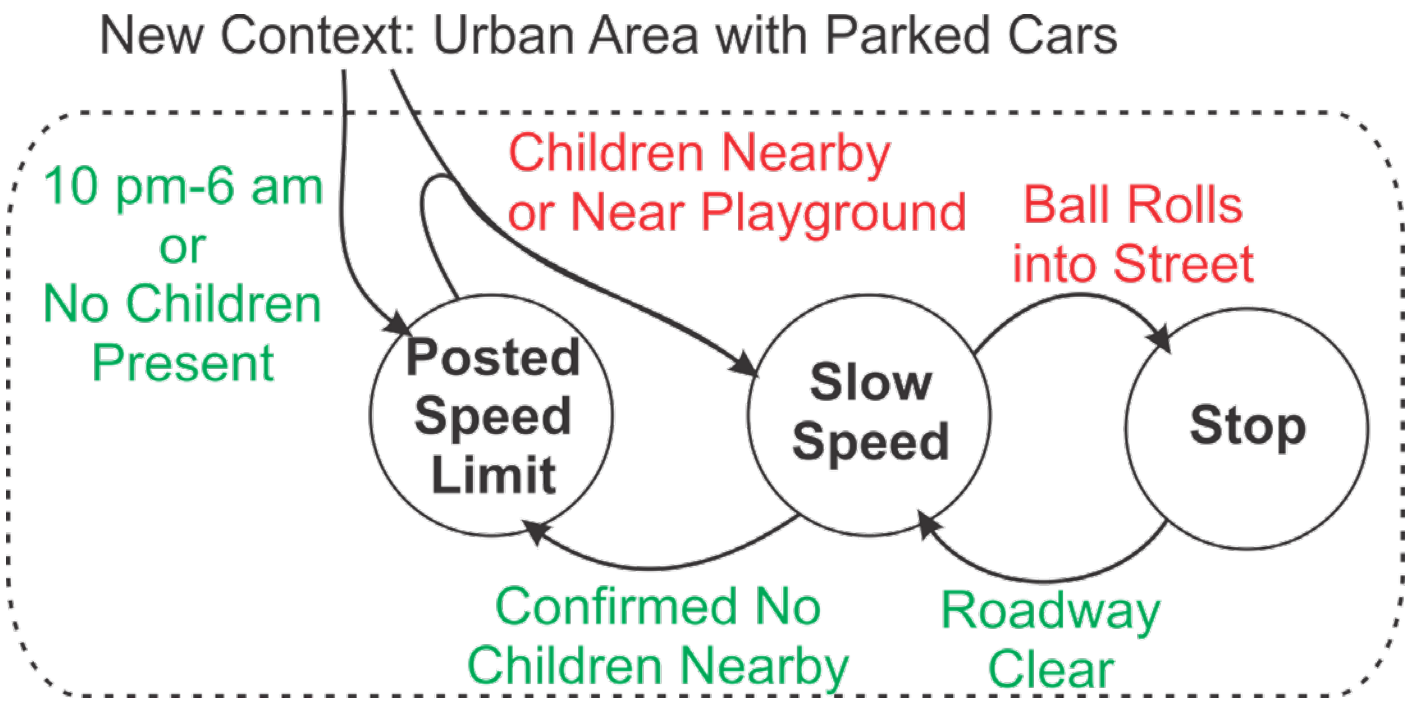

Figure 5. Partial, simplified state transition diagram for urban+children situations.

Similarly, freezing temperatures and recent precipitation should likely trigger a μODD which encodes the likelihood of scattered ice patches even if no ice has been observed on the road. Approaching a bridge in near-freezing weather should cause a transition to a μODD that can accommodate a "Caution: bridge freezes before road surface" scenario even if there is no direct information about the bridge road surface conditions.

An additional benefit of this approach is that it enables a phased deployment approach of creating a baseline vehicle that can operate in a broad range of conditions in a very cautious manner. For example, vehicles might be limited to less than 20 kph in a wide variety of circumstances due to the ability of the vehicle to stop very quickly and the low lethality of pedestrian collisions in such circumstances. Over time, μODDs can be engineered and validated to provide increased permissiveness in specific circumstances, with the vehicle transitioning in and out of those permissive μODDs as conditions change. As this type of system matures its capabilities will increase in response to optimizing common operational situations without sacrificing safety for those situations that have yet to be optimized. This could permit safe deployment more quickly than with a system that depends upon having taken every possible scenario into account to attain safety.

It is recognized that a fully Bayesian approach would require a mathematical framework for managing transitions. Providing that and fully elaborating a Bayesian network approach is left as future work.

### 5.4 Taking advantage of helpful but not perfectly reliable data

A common response to uncertainty of information is to improve information with external communication "vehicle-to-x" links such as V2V and V2I communications [Arena19]. Incorporating that information as factual data has numerous potential problems, including potential invalidity as well as how to ensure safety when there is no other available other vehicle or infrastructure to provide data. However, with a Bayesian approach to μODD selection there is a natural way to treat this data. External data can be used to influence the prior belief as part of the μODD selection. Even map data, which can be very good but perhaps imperfect, can be incorporated into such an approach.

## 6 Conclusion

An examination of RSS has validated the following distance equation for common situations and augmented that formula to handle a class of edge cases for potential collisions that can happen during a braking event. A significant potential impediment to practical adoption of RSS is providing sufficient permissiveness while ensuring safety in extreme conditions such as icy roads and encountering clusters of outlier sensor data. To arrive at a practicable balance between safety and permissiveness, further engagement with government and industry standards organizations is recommended.

Rather than making everything probabilistic to handle uncertainty, we propose dividing an RSS-style safety function's operation into two parts. First, a set of μODDs should be defined to segment the total ODD space into manageable pieces. Each μODD can be proven safe with permissiveness optimized for a defined range of bounded vehicle and environmental values. If those values are violated, the system can switch to a different, more situation-appropriate μODD. Second, a Bayesian-style selection engine can be used to evaluate the prior probability of a set of conditions being true, and then selecting a μODD based on that belief. We believe that this approach will not only enable a reasonable deployment of RSS, but also provides a promising framework for dealing with situations such as how to moderate speed in areas that might have potential children at play and other context-dependent safety relevant vehicle behaviors.

This research was supported by Intel.

## Appendix: Detailed Analysis

The analysis is broken up into two parts based on the situation at the time of a collision if following distance is violated: (1) the rear vehicle stops at a location past the stopping point of the front vehicle; and (2) the rear vehicle stops behind the front vehicle, but passes the position of the front vehicle for some fraction of time during the braking process. (In both cases the “passing” of the front vehicle has to do with the computed position in the absence of a collision.) Below is a more detailed derivation.

All cases the original RSS equation avoids having a situation in which the rear vehicle resting position would be in front of the front vehicle:

$$d'_{min} = MAX\left\{0, \left(v_r\rho + \frac{1}{2}\, a_{max,accel}\rho^2 + \frac{(v_r + \rho a_{max,accel})^2}{2a_{min,brake}} - \frac{v_f^2}{2a_{max,brake}}\right)\right\} \quad \text{(A.1)}$$

This minimum holds as a bound for all cases. However, the minimum might be larger to avoid a transient overlap in position that occurs during the stopping process (the special case).

The relevant aspect of special case is an impact that occurs when both vehicles are moving. (If the front vehicle is stopped at time of impact, the RSS equation applies. If the rear vehicle is stopped there can be no impact.)

Avoiding the special case impact requires first that the following distance account for the closing distance d’’ that is consumed during ρ based on the front vehicle decelerating and the rear vehicle accelerating at the same time. Again, vehicles must still be moving during the entire response time ρ to be worse than the original RSS equation.

$$d''_{min} = (v_r - v_f)\rho + \frac{(a_{max,accel} + a_{max,brake})\rho^2}{2} \quad \text{(A.2)}$$

For the special case to result in a collision $d''_{min}$ must be non-negative, meaning that the rear vehicle must be catching up to the front vehicle during the response time.

At the end of the response time the rear vehicle is going faster than or equal to the speed of the front vehicle (this is the special case, by definition). That means there needs to be enough remaining separation in addition to $d''_{min}$ between the front and rear vehicles for the rear vehicle to just barely catch up to, but not touch, the front vehicle as they both brake. This can be determined by solving for the closest point of approach after the response time.

The distance of each vehicle is a function of time t relative to the end of the response time, taking into account velocity changes during the response time. Both vehicles must either be moving or have simultaneously stopped at the closest point of approach due to being at the same speed (see below). The distances they travel at time t are (note that t=0 is set at end of the response time for notational convenience):

$$d_{r(t)} = (v_r + a_{max,accel}\,\rho)t - \frac{a_{min,brake}\,t^2}{2} \quad \text{(A.3)}$$

$$d_{f(t)} = (v_f - a_{max,brake}\,\rho)t - \frac{a_{max,brake}\,t^2}{2} \quad \text{(A.4)}$$

The separation d’’’ between vehicles required to avoid a collision at time t is the amount travelled by the rear vehicle minus the amount travelled by the front vehicle (i.e., the amount of encroachment of the rear vehicle upon the front vehicle since the end of the response time):

$$d''' = \left(v_r + a_{max,accel}\,\rho\right)t - \frac{a_{min,brake}\,t^2}{2} - \left(\left(v_f - a_{max,brake}\rho\right)t - \frac{a_{max,brake}t^2}{2}\right) \quad \text{(A.5)}$$

Finding the maximum value of d''' over time t from the end of the response time to the time at which the rear vehicle has stopped determines the minimum permissible separation to avoid a collision (ironically, this must be given the name $d'''_{min}$ because it is the minimum safe following distance consumed by that phase of the stopping event). In all situations for the special case $d'''_{min}$ will be non-negative.

A constraint on time t is that we want the speeds of two vehicles to be equal. (Consider: if the rear vehicle is faster they are still closing; if the rear vehicle is slower the closest point of approach has already occurred, potentially at t=0, which is the end of the response time.)

The front vehicle's velocity, as a function of time, is:

$$v_{ft} = v_{f0} - a_{max,brake}(\rho + t) \quad \text{(A.6)}$$

This expands to:

$$v_{ft} = v_{f0} - a_{max,brake}\rho - a_{max,brake}t \quad \text{(A.7)}$$

Where $v_{f0}$ is the velocity of the front vehicle at the start of the response time and $v_{ft}$ is the velocity of the front vehicle at time t.

The rear vehicle's velocity, as a function of time, is:

$$v_{rt} = v_{r0} + a_{max,accel}\rho - a_{min,brake}t \quad \text{(A.8)}$$

Where $v_{r0}$ is the velocity of the rear vehicle at the start of the response time and $v_{rt}$ is the velocity of the rear vehicle at time t.

When these velocities are equal, we have:

$$v_{r0} + a_{max,accel}\rho - a_{min,brake}t = v_{f0} - a_{max,brake}\rho - a_{max,brake}t \quad \text{(A.9)}$$

Rearranging to solve for t:

$$(v_{r0} - v_{f0}) + a_{max,accel}\rho + a_{max,brake}\rho = a_{min,brake}t - a_{max,brake}t \quad \text{(A.10)}$$

$$t = \frac{(v_{r0} - v_{f0}) + (a_{max,accel} + a_{max,brake})\rho}{(a_{min,brake} - a_{max,brake})} \quad \text{(A.11)}$$

Note: $a_{min,brake}$ greater than $a_{max,brake}$ is a condition for being in the special case, so the denominator is always strictly greater than zero.

The value for t will be greater than or equal to zero due to the conditions imposed upon the special case. If t=0, $d'''_{min}$ will evaluate to zero, but $d''_{min}$ still applies to avoid a collision during the response time (i.e., for t=0 the closest point of approach is at the exact end of the response time).

Reiterating the requirements for the special case:

- $v_r + a_{max,accel}\rho > v_f - a_{max,brake}\rho$ (Rear vehicle faster than front vehicle at end of response time)
- $a_{min,brake} > a_{max,brake}$ (Rear vehicle braking capability greater than front vehicle braking)

Now we can assemble the pieces to give a comprehensive solution to minimum following distance.

$$d_{min} = \begin{cases} MAX[d'_{min}, (d''_{min} + d'''_{min})] & ;\ case\ 4\ applies \\ d'_{min} & ;\ otherwise\ (Original\ RSS) \end{cases} \qquad \text{(A.12)}$$

As a sanity check on these equations, it can be shown that Equation 12 the special case is equivalent to RSS for the timepoints when the front and rear vehicles reach rest positions. For minimal separation at rest, the timepoint $t_{stop,f}$ is the stopping time of the front vehicle and $t_{stop,r}$ is the stopping time of the rear vehicle after the response time ρ. (The numerators of these equations are the speed at response time ρ, which is used in the computation of $d''_{min}$)

$$t_{stop,f} = \frac{v_f - a_{max,brake}\rho}{a_{max,brake}} \qquad \text{(A.13)}$$

$$t_{stop,r} = \frac{v_r + a_{max,accel}\rho}{a_{min,brake}} \qquad \text{(A.14)}$$

In this minimal separation at rest case, substituting the values in Equations 13 and 14 into Equation 12 produces the original RSS equation.

$$d_{min} = d''_{min} + d'''_{min} \quad \text{evaluated at } t_{stop,f} \text{ and } t_{stop,r} \qquad \text{(A.15)}$$

$$d_{min} = (v_r - v_f)\rho + \frac{(a_{max,accel} + a_{max,brake})\rho^2}{2} + (v_r + a_{max,accel}\,\rho)t_{stop,r} - \frac{a_{min,brake}\, t_{stop,r}^2}{2} - \left((v_f - a_{max,brake}\rho)t_{stop,f} - \frac{a_{max,brake}t_{stop,f}^2}{2}\right) \tag{A.16}$$

$$d_{min} = (v_r - v_f)\rho + \frac{(a_{max,accel} + a_{max,brake})\rho^2}{2} + (v_r + a_{max,accel}\,\rho)\left(\frac{v_r + a_{max,accel}\rho}{a_{min,brake}}\right) - \frac{a_{min,brake}}{2}\left(\frac{v_r + a_{max,accel}\rho}{a_{min,brake}}\right)^2 - \left((v_f - a_{max,brake}\rho)\left(\frac{v_f - a_{max,brake}\rho}{a_{max,brake}}\right) - \frac{a_{max,brake}}{2}\left(\frac{v_f - a_{max,brake}\rho}{a_{max,brake}}\right)^2\right) \tag{A.17}$$

$$d_{min} = v_r\rho - v_f\rho + \frac{a_{max,accel}\rho^2}{2} + \frac{a_{max,brake}\rho^2}{2} + \frac{(v_r + a_{max,accel}\,\rho)^2}{a_{min,brake}} - \frac{a_{min,brake}}{2}\frac{(v_r + a_{max,accel}\rho)^2}{a_{min,brake}^2} - \frac{(v_f - a_{max,brake}\rho)^2}{a_{max,brake}} + \frac{v_f^2}{2a_{max,brake}} - v_f\rho + \frac{a_{max,brake}\rho^2}{2} \tag{A.18}$$

$$d_{min} = v_r\rho + \frac{a_{max,accel}\rho^2}{2} + \frac{a_{max,brake}\rho^2}{2} + \frac{\left(v_r + a_{max,accel}\,\rho\right)^2}{2a_{min,brake}} + \frac{v_f^2}{2a_{max,brake}} - \frac{v_f^2}{a_{max,brake}} + 2v_f\rho - a_{max,brake}\rho^2 - v_f\rho + \frac{a_{max,brake}\rho^2}{2} - v_f\rho \tag{A.19}$$

$$d_{min} = v_r\rho + \frac{a_{max,accel}\rho^2}{2} + \frac{\left(v_r + a_{max,accel}\,\rho\right)^2}{2a_{min,brake}} - \frac{v_f^2}{2a_{max,brake}} + \frac{a_{max,brake}\rho^2}{2} + 2v_f\rho - a_{max,brake}\rho^2 - v_f\rho + \frac{a_{max,brake}\rho^2}{2} - v_f\rho \tag{A.20}$$

$$d_{min} = v_r\rho + \frac{a_{max,accel}\rho^2}{2} + \frac{\left(v_r + a_{max,accel}\,\rho\right)^2}{2a_{min,brake}} - \frac{v_f^2}{2a_{max,brake}} \tag{A.21}$$

Equation A.21 matches the non-zero portion of the original RSS equation, shown in Equation A.1, confirming the sanity check.